\documentclass[10pt,twocolumn,letterpaper]{article}

\usepackage[pagenumbers]{cvpr} 

\usepackage[pdftex]{graphicx}
\usepackage{amsmath}
\usepackage{amssymb}
\usepackage{booktabs}
\usepackage{algorithm,algcompatible,amsmath}
\usepackage{xcolor}
\usepackage{multirow}
\usepackage{xintexpr}

\def\eg{\emph{e.g.~}} 
\def\ie{\emph{i.e.~}} 
 
\def\etc{\emph{etc.~}} 
 
\def\aka{a.k.a.~}

\newcommand{\Lagr}{\mathcal{L}}
\DeclareMathOperator*{\argmin}{argmin} 
\algnewcommand\INPUT{\item[\textbf{Input:}]}%
\algnewcommand\OUTPUT{\item[\textbf{Output:}]}%
\algnewcommand\PARAM{\item[\textbf{Optimizable Parameter:}]}%

\algdef{SE}[SUBALG]{Indent}{EndIndent}{}{\algorithmicend\ }%
\algtext*{Indent}
\algtext*{EndIndent}

\newlength\myindent 
\usepackage[pagebackref,breaklinks,colorlinks]{hyperref}

\usepackage[capitalize]{cleveref}
\crefname{section}{Sec.}{Secs.}
\Crefname{section}{Section}{Sections}
\Crefname{table}{Table}{Tables}
\crefname{table}{Tab.}{Tabs.}

\def\cvprPaperID{8142} 
\def\confName{CVPR}
\def\confYear{2023}

\begin{document}

    \title{ActiveRMAP: Radiance Field for Active Mapping And Planning}
\author{
Huangying Zhan{$^{1,2}$}, Jiyang Zheng{$^{3,4}$}, Yi Xu{$^{2}$}, Ian Reid{$^{1}$}, Hamid Rezatofighi{$^{3}$}\\
	$^{1}$The University of Adelaide 
	$^{2}$InnoPeak Technology, Inc.\\
	$^{3}$Monash University
        $^{4}$Australian National University
}
\maketitle

\begin{abstract}
A high-quality 3D reconstruction of a scene from a collection of 2D images can be achieved through offline/online mapping methods. In this paper, we explore active mapping from the perspective of implicit representations, which have recently produced compelling results in a variety of applications. One of the most popular implicit representations – Neural Radiance Field (NeRF), first demonstrated photorealistic rendering results using multi-layer perceptrons, with promising offline 3D reconstruction as a by-product of the radiance field. More recently, researchers also applied this implicit representation for online reconstruction and localization (i.e. implicit SLAM systems). However, the study on using implicit representation for active vision tasks is still very limited. In this paper, we are particularly interested in applying the neural radiance field for active mapping and planning problems, which are closely coupled tasks in an active system. We, for the first time, present an RGB-only active vision framework using radiance field representation for active 3D reconstruction and planning in an online manner. Specifically, we formulate this joint task as an iterative dual-stage optimization problem, where we alternatively optimize for the radiance field representation and path planning. Experimental results suggest that the proposed method achieves competitive results compared to other offline methods and outperforms active reconstruction methods using NeRFs.
\end{abstract}
    
    \section{Introduction} \label{sec:introduction}




%

%
\begin{figure}[t!]
        \centering
		\includegraphics[width=1\columnwidth]{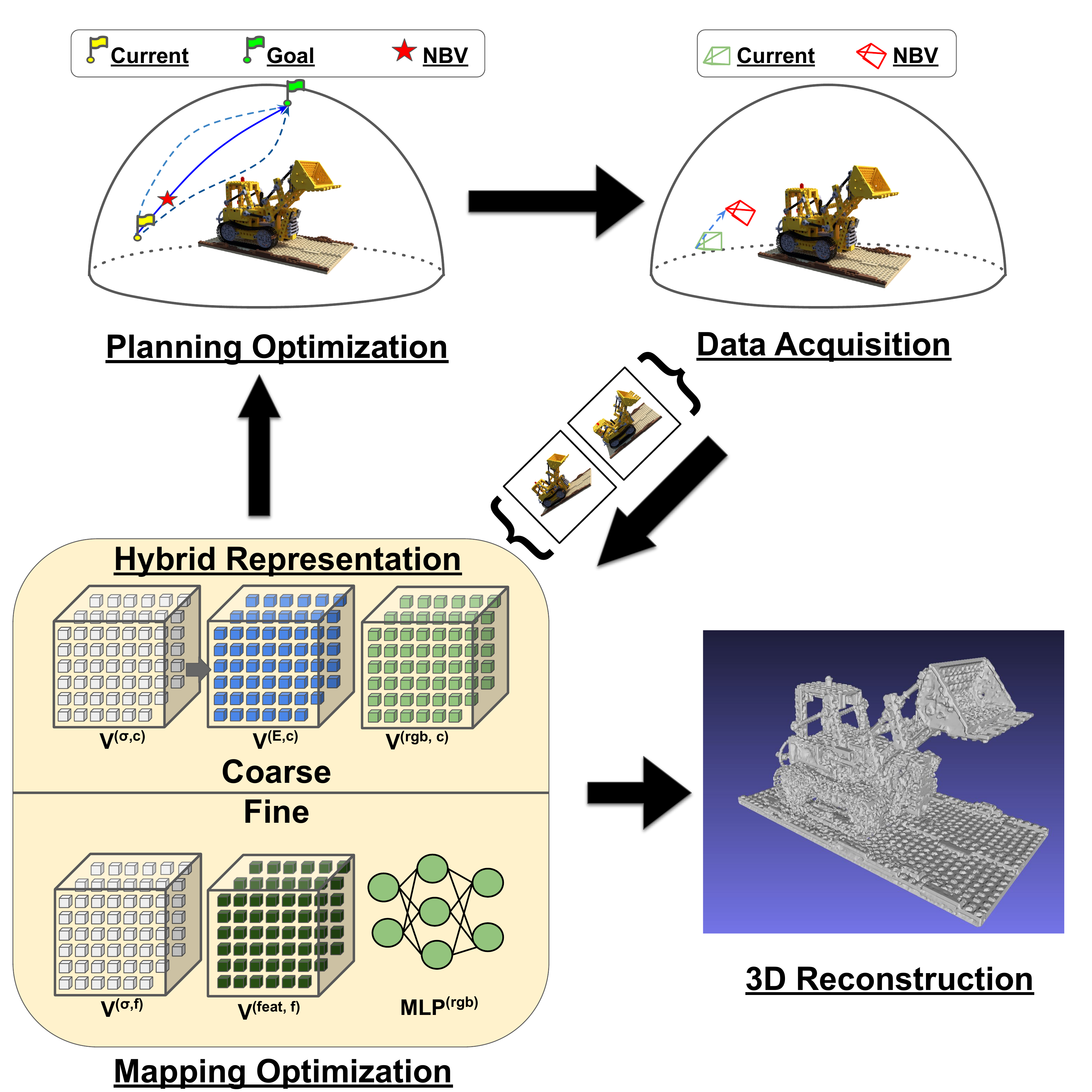}
		\caption{
            \textbf{Overview}. We propose a dual-stage optimization framework for active vision tasks. A hybrid implicit-explicit radiance field is used as the representation of the environment for its speed efficient and remarkable performance on 3D shape representation.
        }
		\label{fig:teaser}
		\vspace{-5pt}
\end{figure}
%

One of the remarkable successes of computer vision research has been to show that it is possible to generate a high-quality 3D reconstruction of a scene from a collection of 2D images or video of the scene. When done offline in batch, this is usually termed structure from motion. When performed \textit{online}, with the 3D reconstruction built incrementally as more visual data is perceived, it is usually termed \emph{Visual Simultaneous Localisation and Mapping} (vSLAM), in which the 3D reconstruction is the map, and the camera poses are localised with respect to the map. vSLAM systems have often been researched and developed with the map as the end product. However in many robotic applications the map is simply a means to perform some other tasks such as planning and navigation; the joint tasks of localisation, mapping, planning, and navigation are collectively known as \textit{Active SLAM}. In this paper we explore Active SLAM from the perspective of powerful, learned, implicit representations of geometry (\ie, NeRF), developing methods that can plan and execute camera paths to optimise the fidelity of the NeRF representation.
Implicit neural representations, especially Neural Radiance Fields (NeRF), have been recently applied in a variety of visual geometric applications, including 3D object reconstruction \cite{park2019deepsdf}, novel view rendering \cite{mildenhall2021nerf, zhang2020nerf++, yu2021pixelnerf, pumarola2021d},
surface reconstruction \cite{li2022bnv, azinovic2022neural},
and generative models \cite{schwarz2020graf, niemeyer2021giraffe}. 
While much of this work assumes well-posed cameras, some work has extended the idea to more general tasks of structure from motion \cite{wang2021nerf--, lin2021barf, chng2022gaussian} and SLAM \cite{sucar2021imap, sun2021neuralrecon}.
The implicit representation is very attractive as an alternative to point clouds or surface meshes for a variety of reasons, including efficient memory usage, continuous and differentiable representation, and natural ways to leverage prior learned information (see, e.g., PixelNeRF \cite{yu2021pixelnerf}). Nevertheless, there remain questions and challenges associated with using this representation within an active or robotic paradigm. To the best of our knowledge, prior work to exploit NeRF for path planning is limited
\cite{adamkiewicz2022vision}, and active reconstruction using NeRF has been considered by three contemporaneous works \cite{ran2022neurar, lee2022uncertainty, pan2022activenerf}. However, \cite{adamkiewicz2022vision} assumes a NeRF model of the scene is pre-trained offline using a collection of images from the environment.
\cite{lee2022uncertainty} has to initialize a coarse NeRF representation by pre-training on a small set of collected images.
The computation time is very slow, which is not desirable for robotic applications.
\cite{ran2022neurar, pan2022activenerf} predict the image reconstruction uncertainty as the criterion for next-best-view selection. 
However, image reconstruction quality is not always a good proxy for geometric reconstruction fidelity.
Moreover, this method requires additional depth supervision to achieve good reconstruction. 

To overcome the above-mentioned drawbacks, 
we propose an iterative dual-stage optimization framework (\cref{fig:teaser}) for active mapping and planning, which operates in an online fashion, \textit{without} pre-training a NeRF model. 
Specifically, the mapping stage involves the optimization of radiance field representation given a set of observed images to date;
the planning stage involves global and local policies for next-best view selection with a multi-objective function that considers obstacle avoidance, geometric information gain, and path efficiency. 
We make the following contributions in creating this framework:
\begin{itemize}
	\setlength{\itemsep}{0pt}
	\setlength{\parskip}{0pt}
	\setlength{\parsep}{0pt}
    \item We propose an iterative dual-stage optimization framework for active reconstruction and planning, which 
    (1) operates in an \textit{online} fashion, without the need of pre-training a NeRF model;
    (2) only requires color images as the visual observation, without the need of a depth sensor, which is usually required in active 3D reconstruction methods.
    \item We propose a new multi-objective loss function for path planning, which allows the agent to plan a path with the consideration of obstacle avoidance and geometric information gain.
    \item We propose a differentiable geometric information gain criterion that considers termination uncertainty (entropy) for next-best view selection. 
    \item We showcase the use of the proposed framework in active 3D reconstruction. 
    Extensive experiments on synthetic and real-world datasets have been performed to show the effectiveness and efficiency of the proposed method.
\end{itemize}

    \section{Related Work} \label{sec:related_work}
\emph{\bf Active Mapping/SLAM: }
%
An autonomous robotic system requires the robot to form a model of the environment, which would require four essential abilities -- including
localization, mapping, planning, and motion control \cite{siegwart2011introduction}.
While the former involves estimating the robot's state, 
creating a representation of the environment, \eg a map, 
planning a path to safely achieve a goal location,
the latter aims to control the movements of the robot according to the planned path.
Localization, mapping, and planning are usually investigated solely or in combination, which results in multiple research areas, such as 
visual odometry \cite{scaramuzza2011visual, zhan2020visual},
stereo matching \cite{sun2003stereo, hirschmuller2005accurate},
multi-view stereo \cite{seitz2006comparison, yao2018mvsnet, liu2022planemvs},
structure-from-motion (SfM) \cite{schonberger2016structure}, 
simultaneous localization and mapping (SLAM) \cite{thrun2002probabilistic, durrant2006simultaneous, davison2007monoslam, cadena2016past},
and active localization/mapping/SLAM.
While SfM is an \textit{offline} method that reconstructs the 3D structure from a set of collected images,
SLAM is an \textit{online} method that incrementally builds the map of an environment while at the same time localizing the robot within the environment.
However, SLAM is still a passive method that requires human operators to scan the environment with a sensor.

It is not concerned with planning, which guides the navigation process.
In contrast, \textit{active} methods \cite{bajcsy1988active, cowan1988automatic, aloimonos1988active} consider planning problems in the loop.
For the cases that aim to improve localization, mapping, and SLAM, the problems are referred to as active localization, active mapping, and active SLAM, respectively.
Active mapping, \aka active reconstruction or next best view (NBV), is a long-standing problem \cite{connolly1985determination}, which aims to search for the optimal movements to create the best possible representation of an environment.
With the assumption of perfect localization, this problem has been primarily addressed to reconstruct scenes and objects of interest from multiple viewpoints \cite{maver1993occlusions, pito1999solution, kriegel2015efficient, isler2016information, delmerico2018comparison, peralta2020next}. 
In this paper, we are interested in the problem that the map of the environment is unknown.
Active localization aims to improve the robot's pose estimation assuming the map is known. 
We refer readers to \cite{fox1998active, borghi1998minimum, mostegel2014active, xie2020survey} for more relevant works.
Active SLAM unifies the three active vision problems (localization, mapping, and planning). 
It allows robots to autonomously perform localization and mapping to reduce the uncertainty of its localization and the representation of the environment. 
Before \cite{davison2002simultaneous} coined the term -- Active SLAM --, this problem has been investigated under different names, majorly as known as exploration problems \cite{thrun1991active, feder1999adaptive, bourgault2002information, makarenko2002experiment, stachniss2004exploration, newman2003autonomous, stachniss2009robotic}.
We refer readers to the survey papers \cite{cadena2016past, lluvia2021active, placed2022survey} for more relevant works, development of the field, and the comparisons between prior works.
In this work, we are more interested in two aspects of the active vision problems -- mapping and planning.

\begin{figure*}[th!]
        \centering
		\includegraphics[width=0.8\textwidth]{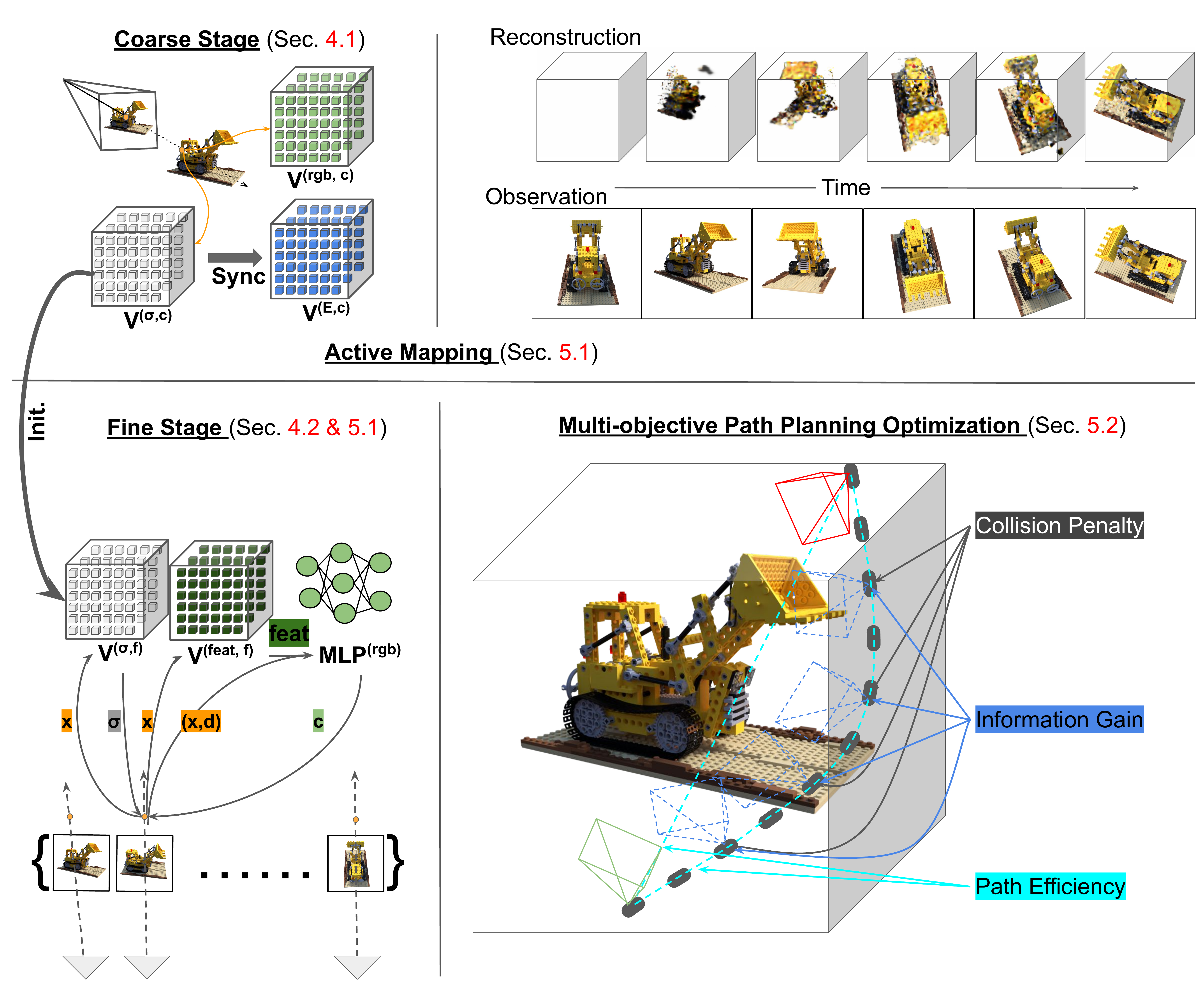}
		\caption{
            \textbf{ActiveRMAP framework}.
            \textbf{Coarse Mapping:} 
            A coarse radiance field representation is optimized with rendering as supervision.
            \textbf{Planning: }
            Path planning is formulated as an optimization problem that considers collision avoidance and information gain of sampled viewpoints on the predicted trajectory, and path efficiency to reach a goal location.
            \textbf{Iterative framework: }
            Conditioned on the coarse model, the planning module obtains the next-best view from the optimized trajectory.
            Thus new observations are acquired incrementally and provided for training the coarse model. 
            \textbf{Fine Mapping: }
            Once the coarse mapping is completed,
            the optimal coarse model is used for initializing the fine model.
            Refinement is performed in the Fine Stage.
        }
		\label{fig:framework}
		\vspace{-10pt}
\end{figure*}
\emph{\bf Neural Radiance Fields:}
Implicit neural representations, especially Neural Radiance Field (NeRF) \cite{mildenhall2021nerf}, have been 
proposed as a new representation of the scene.
NeRF represents the scene as continuous neural radiance fields using multi-layer perceptrons (MLPs), and it can be simply trained by comparing a set of rendered images with well-posed images. 
Despite the compelling properties and results demonstrated by NeRFs, training implicit NeRF can be time-consuming and usually takes about a day for a simple scene, due to the nature of volume rendering \cite{kajiya1984ray} that requires querying on a significant number of sample points in order to render an image.
Though there are some recent works that aim to increase the training and/or inference speed of NeRF \cite{Reiser2021ICCV, deng2022depth, liu2021neulf}, these completely implicit representations are still not fast enough for real-time applications.
More recently, some works \cite{SunSC22dvgo, yu_and_fridovichkeil2021plenoxels, mueller2022instant, Chen2022tensorf} have demonstrated super fast radiance fields with a hybrid representation, \ie implicit for light field and explicit for density field.

NeRF first demonstrated its remarkable ability in 
novel view rendering \cite{mildenhall2021nerf, zhang2020nerf++, yu2021pixelnerf, pumarola2021d}.
Meanwhile, compelling results have been demonstrated for
3D object reconstruction \cite{park2019deepsdf, mildenhall2021nerf}, 
surface reconstruction \cite{li2022bnv, azinovic2022neural},
generative models \cite{schwarz2020graf, niemeyer2021giraffe},
Structure-from-Motion \cite{wang2021nerf--, lin2021barf, chng2022gaussian},
SLAM \cite{sucar2021imap, sun2021neuralrecon},
\etc 
Though implicit representations have been employed in various applications, the study of active vision problems is still very limited. 

\emph{\bf NeRF + Active Vision: }
To our best knowledge, a prior work \cite{adamkiewicz2022vision} shows the use of NeRF for path planning, and three concurrent works \cite{ran2022neurar, lee2022uncertainty} show the use of NeRF for active reconstruction.
\cite{adamkiewicz2022vision} tackles the navigation problem using NeRF representation.
It assumes a NeRF model of the scene is pre-trained offline using a collection of images from the environment.
Thus, an optimal path is searched in the \textit{NeRF map} by minimizing the collision probability of the path extracted from the NeRF representation.
\cite{lee2022uncertainty, pan2022activenerf, ran2022neurar} tackle the active mapping problem by finding the next best view (NBV) to optimize the NeRF representation. 
\cite{lee2022uncertainty} proposes to use ray-based uncertainty as a criterion for selecting NBV, but a coarse NeRF representation has to be initialized by pre-training on a small set of collected images first.
\cite{ran2022neurar, pan2022activenerf} predict the image reconstruction uncertainty as the criterion for NBV selection. 
However, image rendering quality does not directly reflect the geometric reconstruction quality. 
Moreover, \cite{ran2022neurar} requires additional depth supervision to achieve good reconstruction. 
More importantly, these methods inherent the speed drawback from implicit NeRF representations. 
The computation time is very slow, which is not desirable for robotic applications.
In this work, we propose an RGB-only active vision framework based on a hybrid implicit-explicit representation, which actively and incrementally builds a representation of the scene without the need for pre-training.
It has overcome the drawbacks of the existing approaches.
    

\section{Preliminaries}
\label{sec:method:preliminaries}
\subsection{Neural Radiance Field}
NeRF\cite{mildenhall2021nerf} was first proposed for solving novel view synthesis.
The core of NeRF is the use of multi-layer perceptron networks (MLPs) as an implicit representation of the scene.
The MLPs map a 3D position $x$ and a viewing direction $d$ to the corresponding density $\sigma$ and view-dependent color $c$:
\begin{align}
    (\sigma, \textit{e}) &= \text{MLP}_{\Theta_{pos}}^{(pos)}(x); 
    c = \text{MLP}_{\Theta_{rgb}}^{(rgb)}(e, d),
\end{align}
where $\Theta_{pos}$ and $\Theta_{rgb}$ are the learnable MLP parameters, and $\textit{e}$ is an intermediate embedding to help the shallower $\text{MLP}^{(rgb)}$ to learn the color representation.
Given a set of posed images, the training of MLPs relies on the photometric loss between the observed pixel colors $C(r)$ in posed images and the rendered pixel colors $\hat{C}(r)$ in the images viewed from the camera poses , $\Lagr_{photo} = \frac{1}{|R|} \sum_{r \in R} || C(r) - \hat{C}(r) ||$,
%
%
where $R$ denotes a set of rays in a sampled mini-batch.
Volume rendering \cite{kajiya1984ray} is employed to render the pixel colors by casting a ray $r$ from the camera center through the pixel.
Volume rendering approximates light radiance by integrating the radiance along the ray. 
$K$ points are sampled along the ray between pre-defined near and far planes.
After querying for their densities and colors ${(\sigma_i, c_i)}_{i=1}^{K}$ via MLPs, the results are accumulated into a single color with the use of quadrature approximation \cite{max1995optical} as,
\begin{align}
    \hat{C}(r) &= \sum_{i=1}^{K} T_i \alpha_i c_i, \\
    \alpha_i &= 1-\text{exp}(-\sigma_i \delta_i)), \label{eqn:alpha} \\ 
    T_i &= \prod_{j=1}^{i-1}(1-\alpha_j),
\end{align}
where $\alpha_i$ is the light termination probability at the point $i$; 
$T_i$ is the accumulated transmittance from the near plane to point $i$;
$\delta_i$ is the distance between adjacent sampled points.

\subsection{Direct Voxel Grid Optimization} \label{sec:dvgo}
Training a NeRF with the method presented above can be time-consuming and usually takes hours for a simple scene, 
due to the querying cost of volume rendering on a significant number of points.
More recently, some works \cite{SunSC22dvgo, yu_and_fridovichkeil2021plenoxels, mueller2022instant, Chen2022tensorf} have demonstrated super fast radiance fields with a hybrid representation, \ie implicit for color inference and explicit for density inference.
With the potential for real-time applications, we choose DVGO\cite{SunSC22dvgo} among these works as the hybrid representation to use in this paper for the following reasons:
(1) a grid-based explicit representation for the density field, which is efficient for computing volumetric information gain in the later stage;
(2) the dual-stage design allows a fast convergence in the coarse stage and a high-quality reconstruction in the fine stage.
This flexibility in the level of detail suits different levels of active vision task requirements;
(3) competitive results and speed when compared to other hybrid representations.
However, note that our proposed method can be adapted to the other NeRF methods and is not limited to DVGO only.

We introduce the core of DVGO here and refer readers to \cite{SunSC22dvgo} for more details.
DVGO speeds up NeRF by replacing the density MLP with voxel grids.
In the coarse stage, 
two volumes store densities and colors explicitly, represented by $\textbf{V}^{(\sigma, c)}$ and $\textbf{V}^{(rgb, c)}$, respectively.
A query of any 3D point $x$ for density $\sigma^{(c)}$ and color $c^{(c)}$ in the coarse stage is efficient with an interpolation:
\begin{align}
    \sigma^{(c)} &= \text{softplus}(\ddot{\sigma}^{(c)}) 
                = \text{log}(1+\text{exp}(\ddot{\sigma}^{(c)} + b) \\
    \ddot{\sigma}^{(c)} &=  \text{interp}(x, \textbf{V}^{(\sigma, c)})); 
    c^{(c)} = \text{interp}(x, \textbf{V}^{(rgb, c)}),
\end{align}
where shift $b$ is a hyperparameter. 
The density volume stores a raw density $\ddot{\sigma}^{(c)}$, which is further activated via a $\text{softplus}$ operator for the final density.
Note that \cite{SunSC22dvgo} initializes $\textbf{V}^{(\sigma, c)}$ with zero raw density to prevent suboptimal geometry. 

In the fine stage, 
DVGO refines a higher-resolution density voxel grid $\textbf{V}^{(\sigma, f)}$ initialized with the optimized coarse geometry $\textbf{V}^{(\sigma, c)}$. 
The view-invariant $\textbf{V}^{(rgb, c)}$ is replaced with a feature voxel grid $\textbf{V}^{(feat, f)}$ and a shallow MLP for inferring view-dependent color emission.
This hybrid implicit-explicit  representation provides a good balance between speed and quality. 
Finally, queries of 3D points $x$ and viewing-direction $d$ at the fine stage are formulated as,
\begin{align}
    \ddot{\sigma}^{(f)} &=  \text{interp}(x, \textbf{V}^{(\sigma, f)})) \\
    c^{(f)} &= \text{MLP}_{\Theta_{rgb}}^{(rgb)} \left(\text{interp}(x, \textbf{V}^{(feat, f)}, c)), x, d\right).
\end{align}
The major supervision for training DVGO models at the coarse and fine stages is the photometric loss. 
In the following context, we denote the DVGO models at the coarse and fine stage as $\textbf{M}^{(c)}$ and $\textbf{M}^{(f)}$ respectively.


\section{ActiveRMAP}
\label{sec:method:active_nerf}
In this section, 
we present a dual-stage optimization framework, ActiveRMAP (\cref{fig:framework}), for active vision tasks, which includes a mapping stage and a planning stage under an assumption of perfect localization and execution.
We first present the mapping and planning optimization in \cref{sec:map_opt} and \cref{sec:method:planning}. 
Finally, we introduce the application in Active 3D Reconstruction task in \cref{sec:method:reconstruction}.

\subsection{Mapping Optimization} \label{sec:map_opt}
Differing from most NeRF methods, which collect a set of images of the scene and thus optimize the radiance field \textit{offline}, we create an \textit{online} system that acquires and processes the observed images incrementally.

\emph{\bf Coarse mapping: }
Given an observed image $I_t$ at time-\textit{t}, we first store $I_t$ to a database $\{I_i\}$.
Since the database size expands as the agent explores the environment, it is not practical to train the model with all the pixels at each iteration.
Therefore, we random sample a set of pixels from  $\{I_i\}_{i=0}^{t}$ for training $\textbf{M}^{(c)}_{t}$ at times-\textit{t}.
Note that rendering a pixel requires sampling points along the ray, which is time-consuming.
Instead of re-sampling rays at each iteration, we only sample rays once at the beginning of the optimization until a new observation is acquired.

\emph{\bf Fine mapping: }
Once the agent has completed exploring the environment, we can perform fine reconstruction if a high-quality reconstruction is required.

\subsection{Planning Optimization} \label{sec:method:planning}
One of our major contributions is the proposed planning module, which considers collision avoidance, information gain, and path efficiency via a multi-objective optimization formulation.
Given the current agent state (pose) $\textbf{s}_{t}$, a goal state $\textbf{s}_{g}$, and the DVGO coarse model $\textbf{M}^{(c)}_{t}$ at times-\textit{t}, 
we aim to generate a trajectory that yields various task-dependent objectives.
The trajectory is defined as a sequence of agent states 
$\textbf{S} = (\textbf{s}_t, \textbf{s}_{t+1} ..., \textbf{s}_{t+n}, \textbf{s}_g)$, where $\textbf{s}_i$ can be lie-algebra 
or spherical coordinates representation 
depending on the desirable representation in the task.
This problem of finding the optimal trajectory $S^*$ can therefore be expressed as,
\begin{align}
    \textbf{S}^* &= \argmin_{\textbf{S}} \sum_{i=t+1}^{t+n} \Lagr_{plan}(\textbf{M}^{(c)}_{t}, \textbf{s}_i) \\
    \Lagr_{plan} &=  \lambda_{cp}\Lagr_{cp} + \lambda_{oc}\Lagr_{ig} + \lambda_{oc}\Lagr_{pe},
\end{align}
where 
$\Lagr_{plan}$ is the overall planning objective, 
$\Lagr_{cp}$ is the collision penalty objective, 
$\Lagr_{ig}$ is the information gain objective, 
$\Lagr_{pe}$ is the path efficiency objective, 
and $[\lambda_{cp}, \lambda_{ig}, \lambda_{pe}]$ are the weighting terms for each objective.
Once the trajectory is optimized, we choose the furthest state $\textbf{s}^*_i$ that fulfills motion constraints as the next state.
We consider a maximum translation (0.5m) and a maximum rotation angle (10) as our motion constraints.

%

\emph{\bf B\'ezier Trajectory: }
Though we aim to find an optimal sequence of agent states, we do not simply make the agent states the optimizable parameters as this approach can cause infeasible agent motions such as teleporting between distanced viewpoints.
To solve this issue, we use parametric curves for trajectory formation.
Specifically, we adopt the Quadratic B\'ezier curve to create a set of discrete waypoints that define a smooth, continuous trajectory.
For the Quadratic B\'ezier curve, the discrete points can be defined as 
$\textbf{s}_i = (1-r)[(1-r) \textbf{s}_t +  t \textbf{s}_c] + 
        r[(1-r) \textbf{s}_c + r \textbf{s}_g]$, 
%
%
where $r = (i - t)/(n+1)$, and $\textbf{s}_c$ is an optimizable B\'ezier control point, which is initialized as the mid-point of $\textbf{s}_0$ and $\textbf{s}_g$. 

\emph{\bf Collision Penalty:}
The density of the environment is represented as a continuous radiance field, which can be used as a means for computing collision penalty.
\cref{eqn:alpha} computes the light termination probability at a point $x$ based on the density at $x$.
\cite{adamkiewicz2022vision} assumes that a 3D point $x$ with high density (termination probability for light) is a strong proxy for the probability of terminating a mass particle.
We can thus minimize the collision penalty by minimizing the density of the sampled waypoints. 
Thus, the collision penalty is formulated as
\begin{align}
    %
    %
    \Lagr_{cp} &= \sum_{i=t+1}^{t+n}
            l_{cp}(x_i) 
            = \sum_{i=t+1}^{t+n} \text{exp}(\sigma^{(c)}(x_i)),
\end{align}
where 
$\sigma^{(c)}(x)$ queries the density at 3D point $x$.

\emph{\bf Uncertainty-based Information Gain:}
Active mapping requires the evaluation of the quality of un-visited viewpoints -- usually termed information gain. 
A challenge is to quantify the quality and define a meaningful criterion for evaluation, which has not been well studied with the implicit representation.
We address this challenge by using termination uncertainty evaluated from each viewpoint.
For each viewpoint $\textbf{s}_i$, we create a frustum originating from the camera center and sample points within the frustum $\textbf{F}_{i}$.
Applying the same assumption of termination probability, we further compute the uncertainty of termination probability for the sampled points. Therefore, the entropy of a point $E_x(x)$ and a state  $E_s(\textbf{s}_i)$, and the information gain objective are represented by,
\begin{align}
    E_x(x) &= - \alpha(x) \text{log} \alpha(x) - 
                \Bar{\alpha}(x) \text{log} \Bar{\alpha}(x) \\
    \Lagr_{ig} &= - \sum_{\textbf{s}_i \in \textbf{S}} E_s(\textbf{s}_i) 
                = - \sum_{\textbf{s}_i \in \textbf{S}} \left ( \sum_{x_i \in \textbf{F}_i} E_x(x_i) \right)
\end{align}
where $\Bar{\alpha}(x) = 1 - \alpha(x)$.
Intuitively, the viewpoint with a higher entropy means higher uncertainty about the termination probability of the associated 3D points.
The entropy can be decreased once the model obtains more information about the 3D points. Thus it is more certain about the density (thus termination probability) at the sampled points.

Note that although we can directly compute $\alpha(x_i)$ from the density volume, it comes with an initialization issue.
DVGO initializes the density volume with zeros to avoid sub-optimal geometry, which means that the scene has nearly zero entropy at the beginning thus it cannot guide the agent to choose next-best view with the highest information gain.
To overcome this issue, we introduce an additional entropy volume $\textbf{V}^{(E, c)}$ represented by $\alpha^{(E,c)}$ in DVGO coarse model $\textbf{M}^{(c)}$.
We initialize $\alpha^{(E,c)}$ with 0.5, which represents the highest entropy. 
For the voxels of $\textbf{V}^{(\sigma, c)}$ that have been updated in the reconstruction stage (\cref{sec:method:reconstruction}), we synchronize $\alpha^{(E,c)}$ with $\alpha^{(\sigma, c)}$.

%

\emph{\bf Path Efficiency:}
For some active vision tasks, \eg visual navigation, we want a trajectory not only to avoid potential obstacles but also with the shortest path length. 
To achieve path efficiency, we regularize the trajectory by minimizing the following objective, 
$\Lagr_{pe} = \frac{1}{(n)}\sum_{i=t+1}^{t+n} || \textbf{s}_i - \hat{\textbf{s}}_i||$, 
%
%
where $\hat{\textbf{s}}_i$ is a reference state generated by linear interpolating $\textbf{s}_0$ and $\textbf{s}_g$.

\subsection{Active 3D Reconstruction}
\label{sec:method:reconstruction}
%

\vspace{-5pt}
\begin{algorithm} 
    \caption{Active 3D Reconstruction}
  \begin{algorithmic}[1]
    \INPUT DVGO coarse model $\textbf{M}^{(c)}$
    \STATE \textbf{Initialization} agent state $s_t=s_{0}$; goal state candidates $\textbf{S}_g$; image database $\{I\}_{i=0}^{0}$; STOP=False
    \STATE \textbf{Coarse Stage: }
        \Indent
        \WHILE{not STOP}
            \STATE \textbf{Global policy}: obtain a goal state $\textbf{s}_g$ from $\textbf{S}_g$
            \STATE \textbf{Local policy}: $\textbf{S}^*\leftarrow$planning optim. (\cref{sec:method:planning})
            \STATE \textbf{Action}: $\textbf{s}_{t+1} \leftarrow \textbf{s}^*$ 
            \STATE \textbf{Observation}: obtain a new image observation $I_t$
            \STATE \textbf{Update database}: $\{I\}_{i=0}^{t+1} \leftarrow \{I\}_{i=0}^{t}$
            \STATE \textbf{Mapping Optimization}: $\textbf{M}^{(c)}_{t+1} \leftarrow \textbf{M}^{(c)}_t$ 
            \STATE \textbf{Check STOP}: check STOP condition
        \ENDWHILE
        \EndIndent
    \STATE \textbf{Fine Stage: }
        \Indent
            \STATE \textbf{Initialization} DVGO coarse model $\textbf{M}^{(c)}_{T}$; image database $\{I\}_{i=0}^{T}$
            \FOR {$k \leftarrow \text{0 to N}$}
                \IF {$k \mod m == 0$}
                    \STATE \textbf{Training Sample}: sample training pixels (rays) from $\{I\}_{i=0}^{T}$
                \ENDIF 
                \STATE \textbf{Optimize DVGO}: refine $\textbf{M}^{(f)}$ 
            \ENDFOR
        \EndIndent
  \end{algorithmic}
\label{alg:active_recon}
\end{algorithm} 

%
In this section, we present an application of our proposed dual-stage active vision framework for the Active 3D Reconstruction task.
For Active 3D Reconstruction, a robot is tasked with autonomously reconstructing a scene or object of interest. 
We want the agent to create a high-quality and complete 3D reconstruction autonomously without hitting obstacles in the scene while moving around.
A summary of the algorithm is presented in \cref{alg:active_recon}. 

\emph{\bf Initialization:}
We first initialize a 3D bounding box around the space to be reconstructed.
Thus, a set of goal-state candidates are uniformly sampled from the view space.
Note that the goal candidates are not necessarily located in empty spaces. 
We then initialize a DVGO coarse model and an agent state from the current state.
The initial image observation is appended to the image database.

\emph{\bf Coarse Stage:}
At each time step, path planning involves two steps. 
First, a global policy step selects the next-best-view $\textbf{s}_g^*$ from the goal candidates within the whole space, which provides a temporary destination $\textbf{s}_g$.
To evaluate the goal candidates, we adopt the following criterion that considers density and information gain,
\begin{align}
    \textbf{s}_g^* = \argmin_{\textbf{s}_g \in \textbf{S}_g} 
                    (\lambda_{cp} l_{cp}(x_{\textbf{s}_g}) - \lambda_{ig} E_s(\textbf{s}_g)),
\end{align}
where $x_{\textbf{s}_g}$ is the 3D location of the state $\textbf{s}_g$
Next, we execute the path planning method introduced in \cref{sec:method:planning} to obtain an optimized trajectory and the local next-best view $\textbf{s}^*$.
Once the agent maneuvers to the new location and obtains a new observation,
The image database is updated with the new observation.
Samples are drawn from the image database for optimizing the DVGO model, including synchronizing the entropy volume $\textbf{V}^{(E, c)}$.
At the end of the iteration, we check the early stop condition and stop the exploration process if the stop criteria are met.
Here, we consider two stopping criteria:
(1) the change in the entropy volume is below a threshold; and
(2) the maximum information gain of the feasible goal candidates is below a threshold.

\emph{\bf Fine Stage:}
Once the exploration is completed, we initialize the DVGO fine model with the trained coarse model.
Stochastic training is performed by sampling pixels (rays) from the image database.


    
    \section{Experiments and Results} \label{sec:exp_results}



%
\begin{table}[t!]
    \begin{subtable}[h]{1\columnwidth}
        \centering
            \resizebox{1\columnwidth}{!}{%
            \begin{tabular}{ l |  c  c c | c c c } 
                    Methods &
                    PSNR $\uparrow$ & 
                    SSIM $\uparrow$ & 
                    LPIPS $\downarrow$ & 
                    PSNR $\uparrow$ & 
                    SSIM $\uparrow$ & 
                    LPIPS $\downarrow$ \\
                    \hline
                    \hline
                     & \multicolumn{3}{c | }{Discrete (local), 10 views}  & \multicolumn{3}{c}{Discrete (local), 20 views}  \\ 
                    \hline
                    Random &
                    16.869	& 0.789	 & 0.215 &
                    20.435	& 0.839	 & 0.160 \\
            
                    FVS & 
                    18.683 &	0.805 &	0.189 &
                    23.470 &	0.872 &	0.127 \\
            
                    \textbf{Entropy} &
                    \textbf{19.358} &	\textbf{0.820} &	\textbf{0.172} &
                    \textbf{23.936} & 	\textbf{0.880} &	\textbf{0.118} \\
                    \hline
                    \hline
            
                     & \multicolumn{3}{c | }{Discrete (free), 10 views}  & \multicolumn{3}{c}{Discrete (free), 20 views}  \\ 
                    \hline
                    ActiveNeRF \cite{pan2022activenerf} &
                    20.010 & 0.832 & 0.204 &
                    \textbf{26.240}  & 0.856 & 0.124 \\
                    
            
                    \textbf{Entropy} &
                    \textbf{21.853} &	\textbf{0.846} &	\textbf{0.144} &
                    24.489 &	\textbf{0.888} &	\textbf{0.113} \\
                    \hline
                    \hline
                    
                    \multicolumn{7}{c}{Continuous}\\ 
                    \hline
                    \textbf{Ours (Full)} &
                    \textbf{30.181} & \textbf{0.942} & \textbf{0.063} &
                    -  & - & - \\
                    \hline
            \end{tabular}
            }
           \caption{Synthetic-NeRF (Rendering)}
           \label{tab:nbv_syn_discrete}
    \end{subtable}

    \hfill

    \begin{subtable}[h]{1\columnwidth}
        \centering
            \resizebox{1\columnwidth}{!}{%
            \begin{tabular}{ l |  c  c c c  } 
                    Methods &
                    Accu. $\uparrow$ & 
                    Comp. $\uparrow$ & 
                    F1-Score $\uparrow$ & 
                    Chamfer $\downarrow$ \\
                    \hline
                    \hline
                     & \multicolumn{4}{c }{Discrete (local), 10 views}  \\ 
                    \hline
                    Random &
                    0.315 & 0.348 & 0.323 & 0.070 
                    \\
            
                    FVS & 
                    0.377   & 0.376  & 0.372 & 0.048 
                    \\
            
                    \textbf{Entropy} &
                    \textbf{0.407} & \textbf{0.424} & \textbf{0.412} & \textbf{0.026} 
                    \\

                    \hline

                    & \multicolumn{4}{c}{Discrete (local), 20 views}  \\ 
                    \hline
                    
                    Random &
                    0.446 & 0.441 & 0.441 & 0.027 \\

                     FVS & 
                    0.493   & 0.522  & 0.506 & 0.024 
                    \\

                    \textbf{Entropy} &
                    \textbf{0.506} & \textbf{0.548} & \textbf{0.524} & \textbf{0.021} 
                    \\
                    
                    \hline
                    \hline

                    \multicolumn{5}{c}{Continuous}\\ 
                    \hline
                    \textbf{Ours (Full)} &
                    \textbf{0.604} & \textbf{0.602} & \textbf{0.603} & \textbf{0.015} \\
                    
                    \hline
                    \hline
            \end{tabular}
            }
           \caption{Synthetic-NeRF (Geometry)}
           \label{tab:nerf_syn_geometry}
    \end{subtable}
    
    \hfill
    
    \begin{subtable}[h]{1\columnwidth}
        \centering
            \resizebox{0.8\columnwidth}{!}{%
            \begin{tabular}{ l |  c  c c  } 
                        Methods &
                        PSNR $\uparrow$ & 
                        SSIM $\uparrow$ & 
                        LPIPS $\downarrow$\\
                        \hline
                        \hline
                
                        Random &
                       18.169& 0.827& 0.225\\
                                        
                        FVS & 
                        20.240 & 0.844 &  0.204\\
                                        
                        \textbf{Entropy} & \textbf{21.560} &  \textbf{0.848} &  \textbf{0.200}\\
                        \hline
            \end{tabular}
            }
           \caption{Tanks \& Temple (Rendering)}
           \label{tab:nbv_syn_discrete_tt}
    \end{subtable}
     \caption{
        Next best view policy comparison (Rendering \& Geometry). 
        ``\textbf{Discrete (local), N views}" means the camera is limited to local motion and the view space is limited to the discrete viewpoints in the training set. Up to N views can be used for training.
        ``\textbf{Discrete (free)}" means teleporting between viewpoints in the training set is allowed.   ``\textbf{Continuous}'' means the camera moves within a continuous space.
     }
     \label{tab:nbv_comparison}
     \vspace{-5pt}
\end{table}

\subsection{Datasets and Metrics}
We evaluate our framework on synthetic and real-world datasets.
\textbf{Synthetic-NeRF} \cite{mildenhall2021nerf} includes eight objects with realistic images.
Following \cite{mildenhall2021nerf}, we set the image resolution to $800 \times 800$ pixels. 
The dataset has been divided into 100 training views and 200 novel views for testing.
\textbf{Tanks\&Temples} \cite{Knapitsch2017} is a real-world dataset captured by an inward-facing camera. 
We use a set of 5 scenes following \cite{liu2020neural}. 
The image resolution is $1920 \times 1080$. 
$1/8$ of the images in each scene are used for testing.

\emph{\bf Evaluation}: 
There are two categories of evaluation metrics adopted in this work.
For evaluating rendering performance, we adopt the metrics used in \cite{mildenhall2021nerf}, including 
\textit{PSNR}, \textit{SSIM}, and \textit{LPIPS} \cite{zhang2018unreasonable}.
For reconstruction evaluation, we evaluate the reconstruction quality on a mesh surface where 10,000 points are sampled for evaluation. 
Four metrics are reported, including
(1,2) \textit{Accuracy/Completeness } measures the fraction of predicted/reference points that are closer to the reference/predicted points than a threshold distance, which is set to 1cm;
(3) \textit{F1 score} is the harmonic mean of accuracy and completeness, which quantifies the overall reconstruction quality;
(4) \textit{Chamfer Distance} measures the distance between nearest neighbor correspondences of two point clouds.
Note that we use the DVGO models trained offline with full training set as the reference ground truth for evaluation. 

\subsection{Implementation details}
\emph{\bf View space: }
NeRFs \cite{mildenhall2021nerf, SunSC22dvgo} are trained \textit{offline} with a collection of images. 
However, as we are interested in active vision tasks, we provide images incrementally while training.   
We first divide our experiments into two sets, defined by the motion spaces, namely \textit{discrete space} and \textit{continuous space}.
\textbf{Discrete space: } viewpoints are limited to the viewpoints in the training set. 
This is further divided into \textit{discrete (free-view)} where the camera is allowed to teleport between any viewpoints and \textit{discrete (local)} where the camera is only allowed to move to one of the three nearest viewpoints.
\textbf{Continuous space: } we allow the camera to move freely within a predefined view space. 
To this end, we need to employ a simulator for rendering novel viewpoints. 
Since \textit{offline} DVGO \cite{SunSC22dvgo} has shown compelling rendering and reconstruction results, we employ it for a simulation purpose, which implies that DVGO is our model's performance upper bound. 

For \textbf{Synthetic-NeRF}, 
the view space is defined as a hemisphere, where the discrete experiments are on the surface of the hemisphere while the continuous experiments are within the space.
For \textbf{Tanks\&Temples},
the view space is within a 3D bounding box defined by the cameras in the training set.

\emph{\bf Mapping optimization: }
In the coarse mapping stage, the model is provided with new observations incrementally in the coarse stage, and all the observations are provided in the fine stage.
We sample 8,192 rays from the images to train the coarse models for 5,000 iterations and perform refinement for 20,000 iterations.
New observations are inserted for every 50 iterations in the coarse stage.
Adam optimizer \cite{kingma2014adam} is used with an initial learning rate $10^-1$ for all experiments.
Scene parameters, \eg grid size, are dataset-dependent.
We adopt the settings from \cite{SunSC22dvgo}.

\emph{\bf Planning optimization: }
The path planning optimization is carried out with Adam optimizer \cite{adamkiewicz2022vision}, with a learning rate $10^-1$ and 100 training iterations for each planning.
[$\lambda_{cp}$, $\lambda_{ig}$, $\lambda_{pe}$] are set to be [10, 1, 0.1].
For the early stop condition, we set the thresholds to be $5\times 10^{-5}$.
Applying the parametric B\'ezier curve, we generate 100 samples along the curve for planning optimization.

We ran three times for all sets of the experiment and reported the mean result for each set. 
For each run, we start with a random initial location, which is shared across all sets of the experiment.

\subsection{Is termination entropy a good criterion?}
In \cref{sec:method:planning}, we propose the use of termination uncertainty (entropy) of 3D points observed by a viewpoint to determine the information gain of the viewpoint.
To validate the effectiveness of the criterion, we compare the approach with some baselines and a prior work\cite{pan2022activenerf}.
The comparison is presented in \cref{tab:nbv_comparison}.
\emph{\bf Baseline (Random)}: a viewpoint is selected randomly as the next best view (NBV) from the viewpoint candidates.
\emph{\bf Baseline (FVS)}: furthest view sampling is applied to select the most distanced viewpoint compared with the current observation set as the NBV.
\emph{\bf ActiveNeRF} selects NBV based on image rendering uncertainty.
Few-shot active learning experiments are designed to compare the performance of the approaches.
New observations are added incrementally while training, up to 10/20-views.
We evaluate the learned representations based on rendering and 3D reconstruction quality.
In the cases the camera is limited to local movements, our proposed entropy-based method consistently outperforms other baselines.
Our method shows better performance in most of the metrics when compared to ActiveNeRF \cite{pan2022activenerf}.

\subsection{Active 3D Reconstruction}
\begin{table} [t!]  
    \begin{center}
    \resizebox{0.8\columnwidth}{!}{%
    \begin{tabular}{ l |  c  c c  } 
        Methods &
        PSNR $\uparrow$ & 
        SSIM $\uparrow$ & 
        LPIPS $\downarrow$ \\
        \hline
        \hline
        
        \multicolumn{4}{c}{\textit{offline training}} \\ \hline
        
        NeRF \cite{mildenhall2021nerf} &
        31.01 & 0.947 & 0.081 \\

        Mip-NeRF \cite{barron2021mip} &
        33.09 & 0.961 & 0.043 \\

        FastNeRF \cite{garbin2021fastnerf} &
        29.97 & 0.941 & 0.053 \\
        
        DVGO \cite{SunSC22dvgo} &
        31.82 & 0.955 & 0.055 \\
        \hline
        \hline
        \multicolumn{4}{c}{\textit{online training}} \\
        \hline
        

        Ours &
        30.18 & 0.942 & 0.063 \\

        \hline

    \end{tabular}
    }
    \end{center}
    \caption{
        Quantitative comparison for novel view synthesis on Synthetic-NeRF. \textit{offline} methods trains the model with all collected images while our method is an \textit{online} method that incrementally adds images into training after the model selected the next best view. 
        Moreover, our application scenario is an even harder setting due to a broader NBV searching space. 
        Note that Ours uses images rendered from DVGO for training, thus DVGO is the performance upper bound of Ours.
    }
    \label{table:exp:nvs_benchmark}
    \vspace{-10pt}
\end{table}
\subsection{Rendering evaluation}
We evaluate our proposed active framework for 3D reconstruction based on rendering and reconstruction results.
We compare our method against prior arts on the task of novel view synthesis.
To our best knowledge, we are the first RGB-based work that trains radiance field from images incrementally acquired in a continuous space, without the need for depth data \cite{ran2022neurar} or pre-training \cite{lee2022uncertainty}. 
Though this \textit{online} setting is more difficult than offline training, we still show competitive results when compared to the state-of-the-art methods.

\subsection{Reconstruction evaluation}
\begin{figure}[th!]
        \centering
		\includegraphics[width=1.0\columnwidth]
        {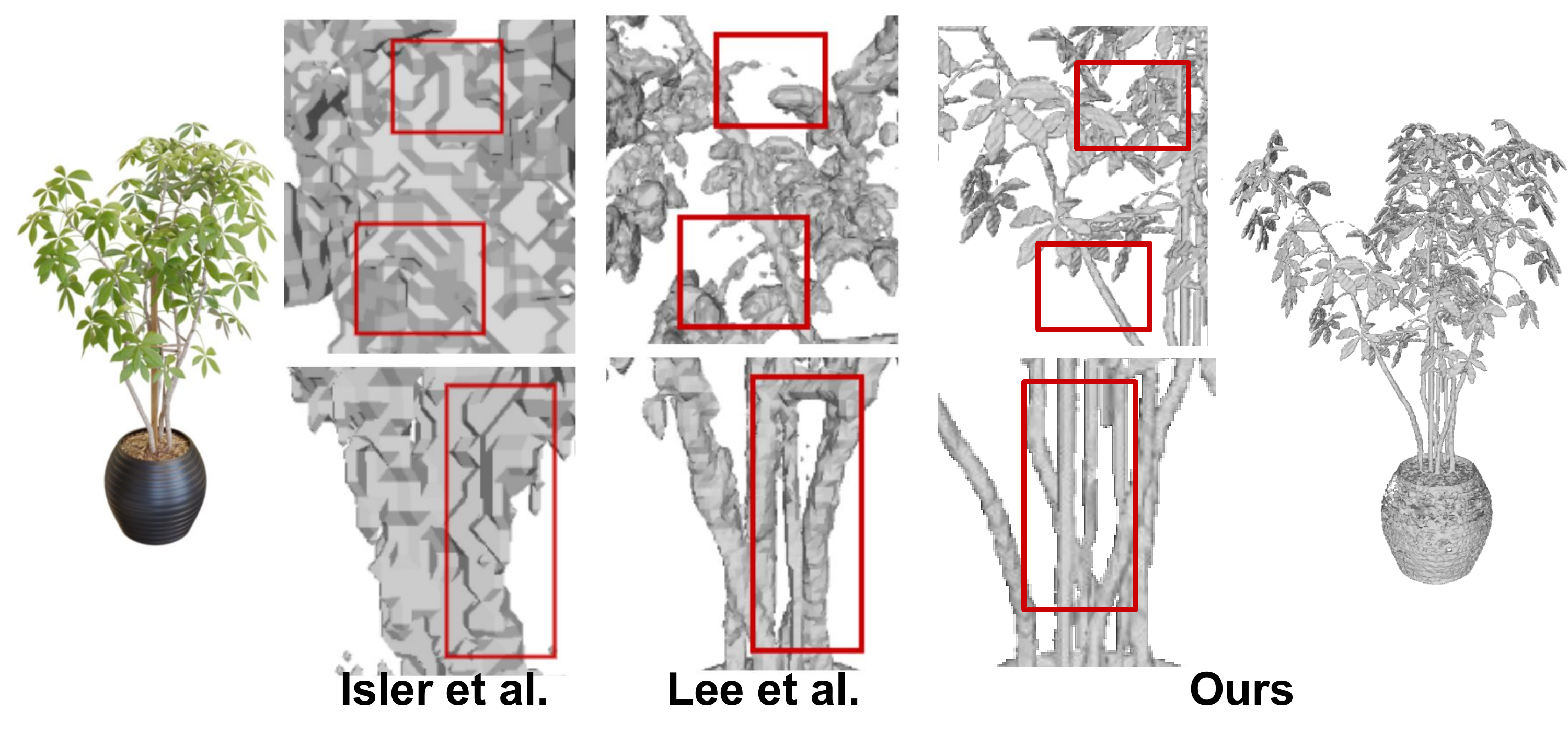}
		\caption{
            Active reconstruction comparison on Synthetic-NeRF (ficus). Our proposed method is capable of generating high-quality 3D models.
        }
		\label{fig:ficus}
		\vspace{-8pt}
\end{figure}
\begin{figure}[t!]
        \centering
		\includegraphics[width=0.8\columnwidth]{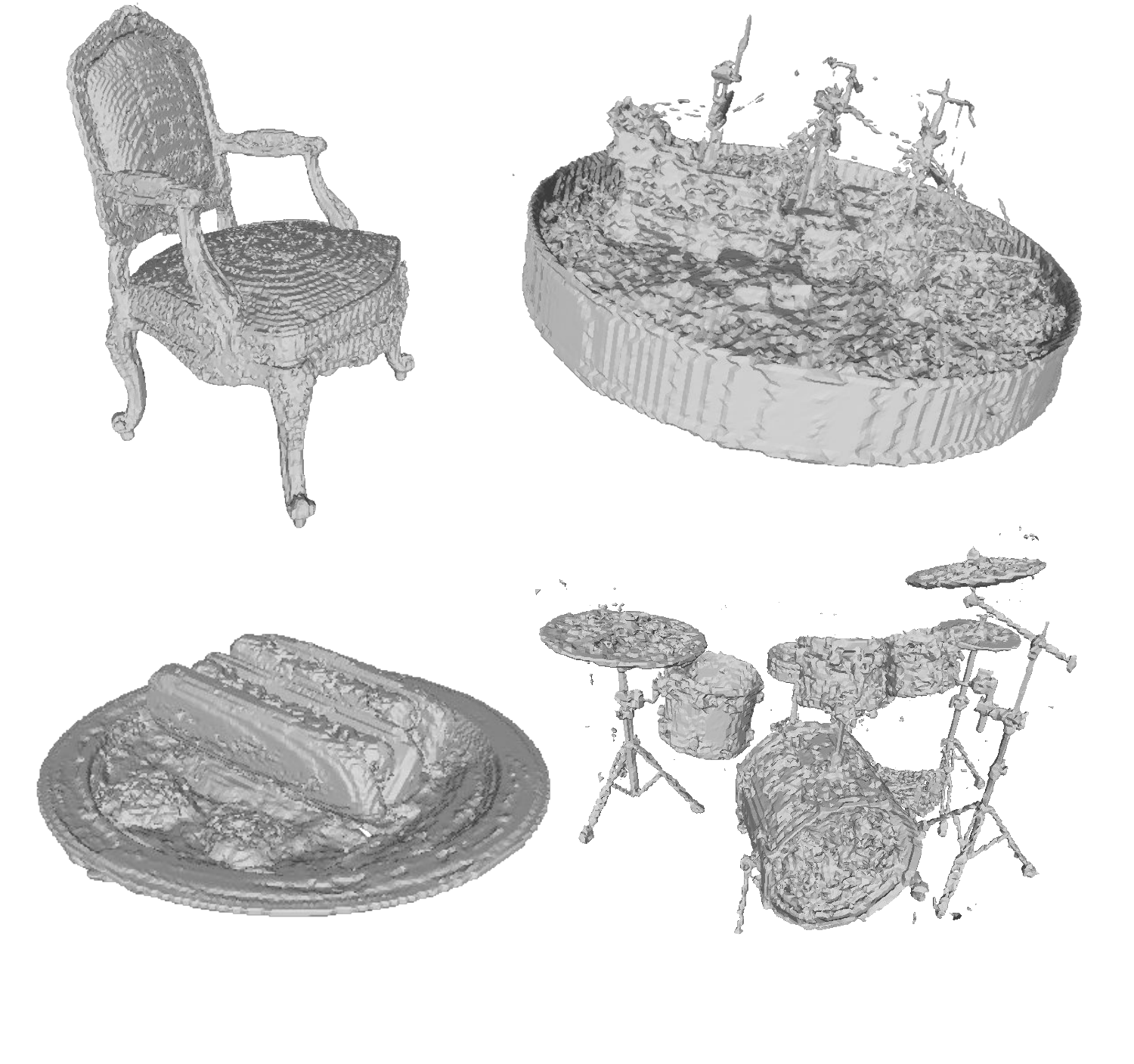}
		\vspace{-2em}
		\caption{
            More 3D reconstruction results on Synthetic-NeRF. Our method has shown good reconstruction on \textit{Chair} and \textit{hotdog} but shows minor noises in difficult scenes like \textit{Ship} and \textit{Drums}. 
        }
		\label{fig:more_eg}
		\vspace{-10pt}
\end{figure}

We present a quantitative comparison of our method against baselines in \cref{tab:nerf_syn_geometry} in \cref{fig:ficus}. 
To compare against prior work, we compare against \cite{isler2016information} and \cite{lee2022uncertainty} qualitatively. 
The results of \cite{isler2016information, lee2022uncertainty} are obtained from \cite{lee2022uncertainty}.
However, as the F1 threshold is not provided in \cite{lee2022uncertainty}, we cannot compare the quantitative results fairly.
From \cref{fig:ficus}, we can see that our methods provide better details among the active reconstruction methods.
More examples are shown in \cref{fig:more_eg}.



    \section{Conclusion} \label{sec:conclusion}
In this work, we tackle the mapping and planning problems in \textit{Active SLAM}, which plays an essential role in autonomous robotic systems.
An iterative dual-stage optimization framework, \textit{ActiveRMAP}, is proposed to perform active mapping and planning with the use of radiance field. 
Unlike time-consuming NeRFs, we employ an efficient implicit-explicit representation in the mapping stage, which allows our framework to have great potential in robotic applications demanding real-time operations.
Photometric difference between observations and volume-rendered images is used as the supervision for mapping optimization.
For the planning stage, we propose a new multi-objective loss function that considers collision avoidance, geometric information gain, and path efficiency into consideration.
Thanks to the continuity property of the radiance field, we formulate the planning part as a fully differentiable optimization problem. 
To our best knowledge, though there are concurrent works \cite{ran2022neurar, lee2022uncertainty} that address the active 3D reconstruction problem using NeRF, we are the first method without the need for pre-training \cite{lee2022uncertainty} and depth supervision \cite{ran2022neurar}.
Our \textit{online} method has shown a competitive result compared to prior \textit{offline} methods and outperforms other active reconstruction methods using NeRFs.

    \newpage
    \section{Overview}
In this supplementary material, we provide more implementation details of ActiveRMAP in \cref{sec:supp:more_imple_details}. 
More results and analysis is provided in \cref{sec:supp:more_results}.

\section{More ActiveRMAP details} \label{sec:supp:more_imple_details}
\subsection{Safe zone modeling}
In the main paper Sec. 4.2., we have introduced a collision penalty for calculating the collision cost based on the continuous radiance field. 
We described the process as minimizing the density of the sampled waypoints. 
However, in practice, we add a safe zone. 
The safe zone is defined by a 3D bounding box centered at the waypoints to simulate the agent body and maintain a safe distance away from high-density regions.
A set of 3D points $\textbf{B}_i$ centered at waypoint state $\textbf{s}_i$ are sampled from the 3D bounding box for computing the penalty.
Thus, the full collision penalty is formulated as
\begin{align}
    l_{cp}(x) &=  \text{exp}(
    \sigma^{(c)}(x)
    ) \\
    \Lagr_{cp} &= \sum_{i=t+1}^{t+n} \left[
            \sum_{x_b \in \textbf{B}_i} 
            l_{cp}(x_b)
    \right],
    %
\end{align}
where 
$\sigma^{(c)}(x)$ queries the density at 3D point $x$.

The safe zone setting should be agent-dependent as it should consider the agent's size.
As this work performs experiments in simulations, we do not have a specific robotic agent.
Thus, we create a 3D bounding box with $1\times1\times1$ unit as the safe zone. 
100 points are sampled uniformly from the bounding box for computing the collision penalty.

\subsection{View space and Sampling}
For \textbf{Synthetic-NeRF}, we define the view space as a hemisphere where the maximum radius is 4. 
We use the spherical coordinate system as the viewpoint representation.
However, the spherical coordinate system only represents the 3D coordinates without rotation. 
Therefore, we constrain the viewpoint to be pointing towards the center of the object of interest.
Sampling is performed when generating waypoints from the trajectory and generating 3D points of interest for information gain calculation.
We sample 100 waypoints uniformly from the B\'ezier trajectory.
For calculating information gain, we sampled 10 viewpoints from the waypoints to reduce the computation cost.
Instead of sampling 3D points from all the rays (pixels) for information gain calculation, we only sample 3D points from $1/100$ of the pixels. 

For \textbf{Tanks \& Temple}, we define the view space as a 3D bounding box whose corners are defined from the raw training set.
The viewpoint state is represented by 6DoF $se3$.
The sampling process is as same as the one described above.




\section{More results} \label{sec:supp:more_results}
\begin{figure*}[th!]
        \centering
		\includegraphics[width=1\textwidth]{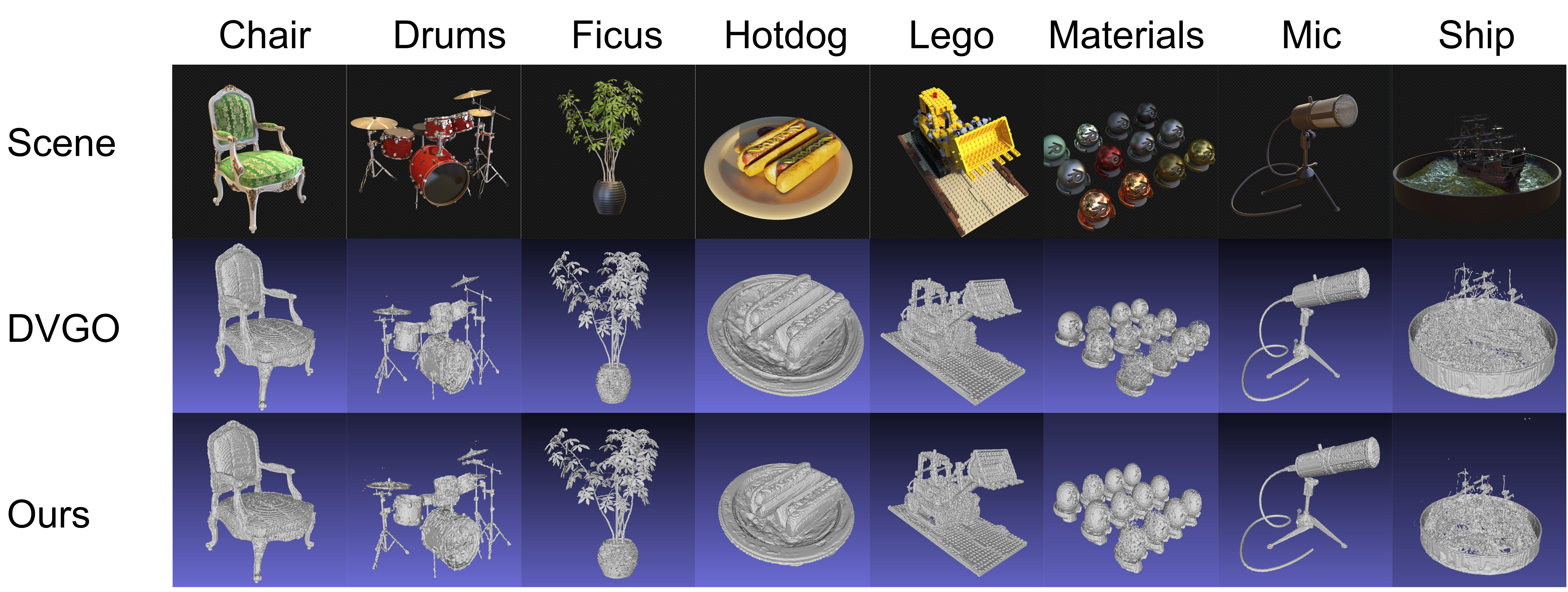}
		\caption{
            NeRF-Synthetic qualitative comparison between our proposed active 3D reconstruction method and DVGO\cite{SunSC22dvgo}.
            Note that, our method uses the images rendered by DVGO as the training images, that implies that DVGO serves as the ``ground-truth'' and the upper bound of our method. 
            Nevertheless, we show a competitive result when compared to DVGO.
        }
		\label{fig:supp:nerf_ours}
\end{figure*}
\begin{figure*}[th!]
        \centering
		\includegraphics[width=1\textwidth]{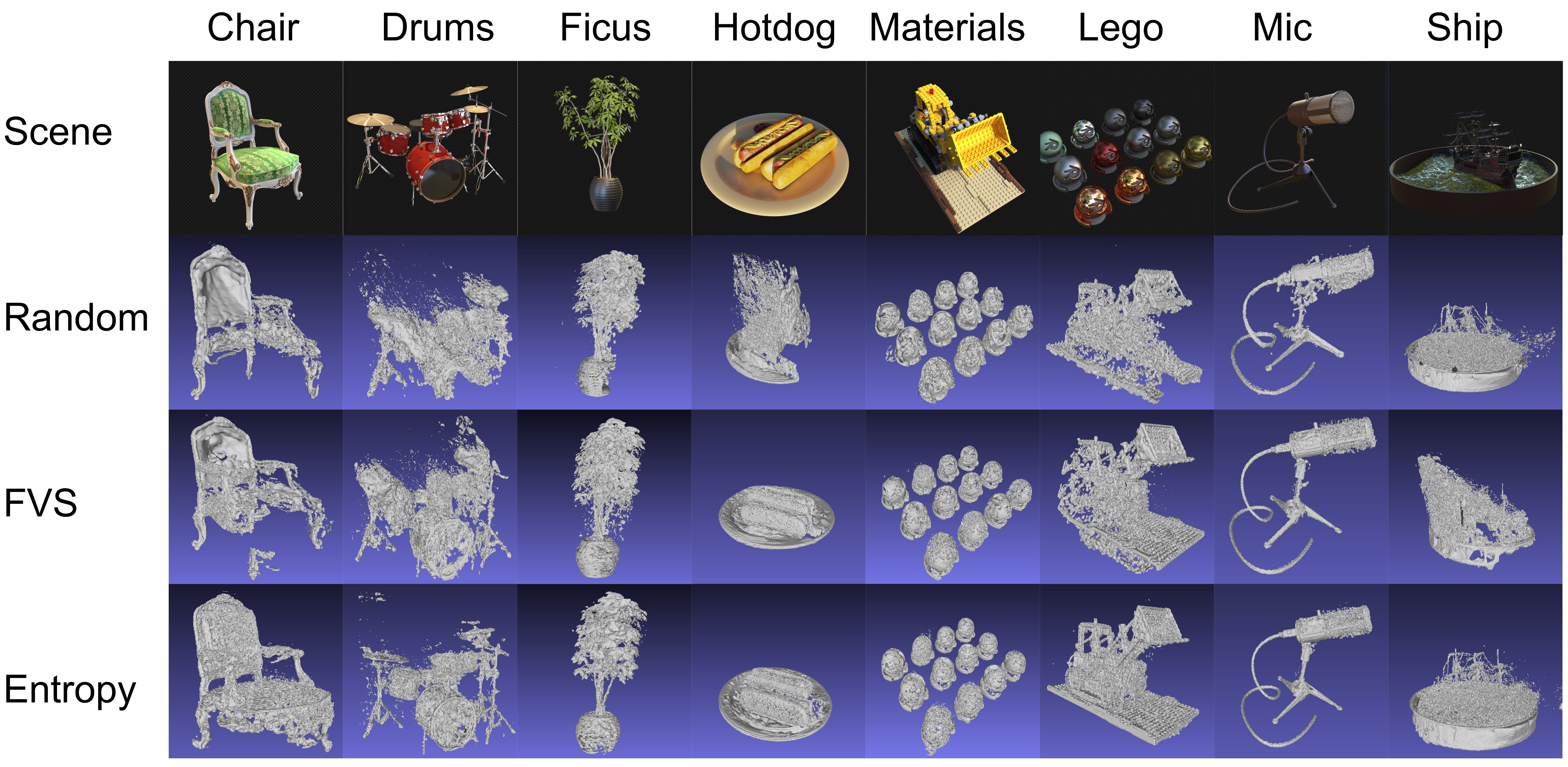}
		\caption{
            NeRF-Synthetic qualitative comparison for termination entropy study.
            We compare the 3D reconstruction model of individual scenes. 
            In this set of experiments, the viewpoints are limited to the viewpoints in the raw training set and the movement is contrained locally, \ie the camera can only moved to one of the three nearest viewpoints.
            The maximum number of viewpoints is 10 in the experiments.
        }
		\label{fig:supp:nerf_discrete_local_10}
\end{figure*}
\begin{figure*}[th!]
        \centering
		\includegraphics[width=1\textwidth]{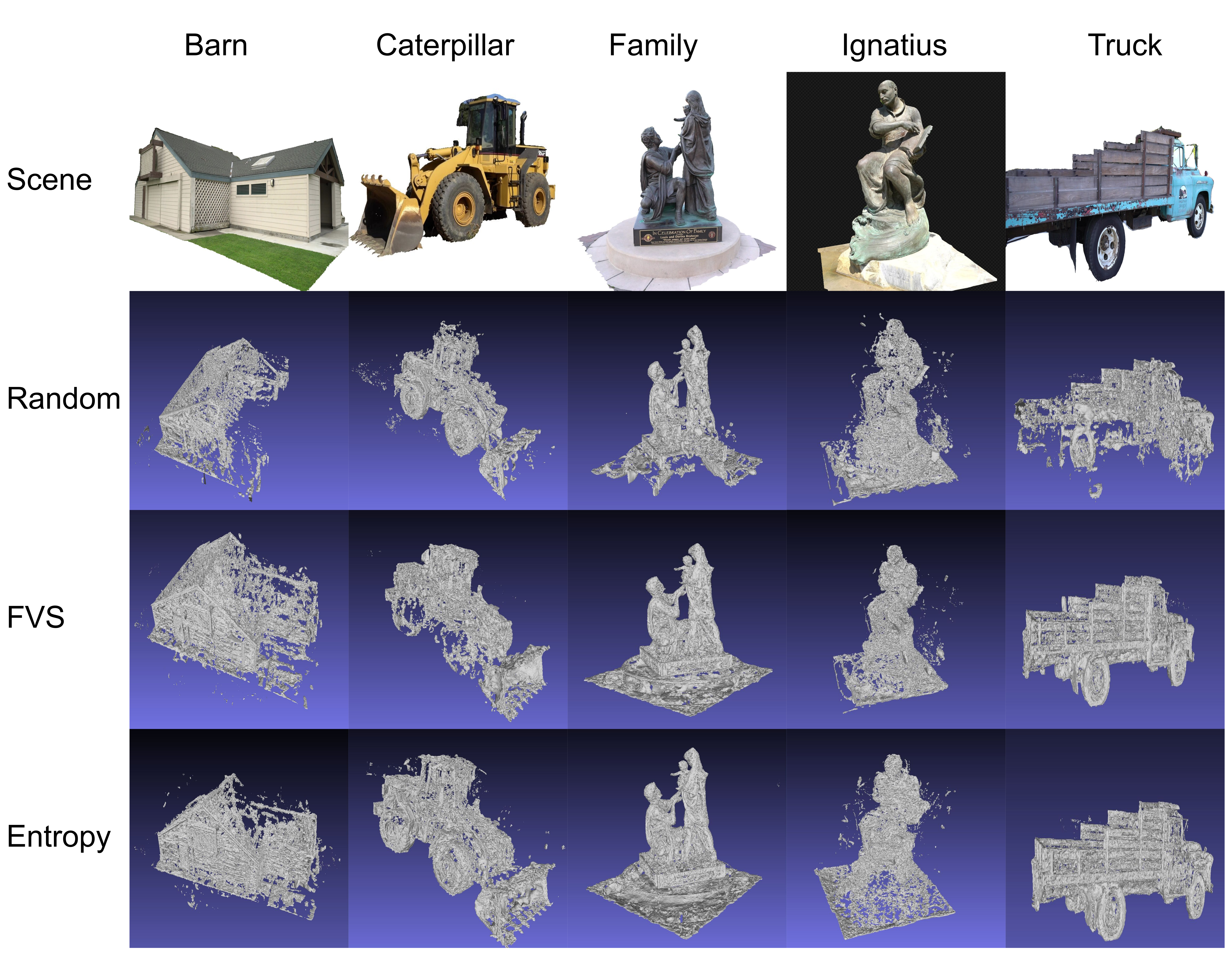}
		\caption{
            Tanks \& Temple qualitative comparison for termination entropy study.
            We compare the 3D reconstruction model of individual scenes. 
            In this set of experiments, the viewpoints are limited to the viewpoints in the raw training set, and the movement is constrained locally, \ie the camera can only move to one of the three nearest viewpoints.
            The maximum number of viewpoints is 50 in the experiments.
        }
		\label{fig:supp:tt_discrete_local_50}
\end{figure*}
\begin{table}[t!]
    \begin{subtable}[h]{1\columnwidth}
        \centering
            \resizebox{1\columnwidth}{!}{%
            \begin{tabular}{ l |  c  c c | c c c c } 
                    Scene &
                    PSNR $\uparrow$ & 
                    SSIM $\uparrow$ & 
                    LPIPS $\downarrow$ & 
                    Accu. $\uparrow$ & 
                    Comp. $\uparrow$ & 
                    F1-Score $\uparrow$ & 
                    Chamfer $\downarrow$ \\
                    \hline
                    \hline
                    chair & 34.100 & 0.976 & 0.027 & - & - & - & - \\
                    drums & 25.400 & 0.930 & 0.079 & - & - & - & - \\
                    ficus & 32.500 & 0.977 & 0.025 & - & - & - & - \\
                    hotdog & 36.700 & 0.980 & 0.035 & - & - & - & - \\ 
                    lego & 34.680 & 0.976 & 0.027 & - & - & - & - \\
                    materials & 29.500 & 0.950 & 0.059 & - & - & - & - \\
                    mic & 33.100 & 0.982 & 0.018 & - & - & - & - \\
                    ship & 28.600 & 0.872 & 0.168 & - & - & - & - \\ 
                    \hline
                    Avg. & 31.823 & 0.955 & 0.055 & - & - & - & - \\
                    \hline
                    \hline
            \end{tabular}
            }
           \caption{DVGO Baseline (\textit{offline training})}
           \label{tab:supp:nerf_baseline}
    \end{subtable}

    \begin{subtable}[h]{1\columnwidth}
        \centering
            \resizebox{1\columnwidth}{!}{%
            \begin{tabular}{ l |  c  c c | c c c c } 
                    Scene &
                    PSNR $\uparrow$ & 
                    SSIM $\uparrow$ & 
                    LPIPS $\downarrow$ & 
                    Accu. $\uparrow$ & 
                    Comp. $\uparrow$ & 
                    F1-Score $\uparrow$ & 
                    Chamfer $\downarrow$ \\
                    \hline
                    \hline
                    chair & 32.770 & 0.968 & 0.037 & 0.639 & 0.637 & 0.638 & 0.011 \\
                    drums & 24.330 & 0.916 & 0.010 & 0.699 & 0.708 & 0.703 & 0.010 \\ 
                    ficus & 31.400 & 0.973 & 0.030 & 0.898 & 0.905 & 0.902 & 0.006 \\ 
                    hotdog & 34.200 & 0.970 & 0.054 & 0.424 & 0.411 & 0.418 & 0.016 \\ 
                    lego & 32.470 & 0.965 & 0.041 & 0.678 & 0.658 & 0.668 & 0.009 \\ 
                    materials & 28.550 & 0.941 & 0.074 & 0.566 & 0.570 & 0.568 & 0.011 \\ 
                    mic & 31.830 & 0.976 & 0.026 & 0.803 & 0.808 & 0.806 & 0.007 \\ 
                    ship & 25.900 & 0.827 & 0.228 & 0.120 & 0.117 & 0.119 & 0.053 \\ 
                    \hline 
                    Avg. & 30.181 & 0.942 & 0.063 & 0.604 & 0.602 & 0.603 & 0.015 \\
                    \hline
                    \hline
            \end{tabular}
            }
           \caption{Ours (Full), continuous space}
           \label{tab:supp:nerf_full}
    \end{subtable}

    \begin{subtable}[h]{1\columnwidth}
        \centering
            \resizebox{1\columnwidth}{!}{%
            \begin{tabular}{ l |  c  c c | c c c c } 
                    Scene &
                    PSNR $\uparrow$ & 
                    SSIM $\uparrow$ & 
                    LPIPS $\downarrow$ & 
                    Accu. $\uparrow$ & 
                    Comp. $\uparrow$ & 
                    F1-Score $\uparrow$ & 
                    Chamfer $\downarrow$ \\
                    \hline
                    \hline
                    chair & 20.800 & 0.860 & 0.138 & 0.391 & 0.403 & 0.397 & 0.021 \\ 
                    drums & 15.200 & 0.759 & 0.242 & 0.347 & 0.407 & 0.375 & 0.026 \\ 
                    ficus & 20.200 & 0.870 & 0.111 & 0.482 & 0.726 & 0.579 & 0.014 \\ 
                    hotdog & 24.500 & 0.907 & 0.114 & 0.329 & 0.339 & 0.334 & 0.024 \\ 
                    lego & 18.960 & 0.814 & 0.174 & 0.434 & 0.404 & 0.419 & 0.021 \\ 
                    materials & 19.600 & 0.833 & 0.148 & 0.522 & 0.482 & 0.501 & 0.013 \\ 
                    mic & 22.100 & 0.906 & 0.088 & 0.624 & 0.519 & 0.566 & 0.012 \\ 
                    ship & 13.500 & 0.608 & 0.361 & 0.129 & 0.115 & 0.121 & 0.073 \\ 
                    \hline
                    Avg. & 19.358 & 0.820 & 0.172 & 0.407 & 0.424 & 0.412 & 0.026 \\ 
                    \hline
                    \hline
            \end{tabular}
            }
           \caption{Ours (Entropy), Discrete (local), 10 views}
           \label{tab:supp:nerf_entropy_10}
    \end{subtable}

    \begin{subtable}[h]{1\columnwidth}
        \centering
            \resizebox{1\columnwidth}{!}{%
            \begin{tabular}{ l |  c  c c | c c c c } 
                    Scene &
                    PSNR $\uparrow$ & 
                    SSIM $\uparrow$ & 
                    LPIPS $\downarrow$ & 
                    Accu. $\uparrow$ & 
                    Comp. $\uparrow$ & 
                    F1-Score $\uparrow$ & 
                    Chamfer $\downarrow$ \\
                    \hline
                    \hline
                    chair & 21.167 & 0.861 & 0.135 & 0.379 & 0.307 & 0.340 & 0.052 \\ 
                    drums & 15.030 & 0.748 & 0.254 & 0.379 & 0.308 & 0.339 & 0.052 \\ 
                    ficus & 19.667 & 0.863 & 0.119 & 0.489 & 0.711 & 0.579 & 0.013 \\ 
                    hotdog & 22.100 & 0.888 & 0.146 & 0.338 & 0.349 & 0.344 & 0.022 \\ 
                    lego & 19.630 & 0.816 & 0.171 & 0.430 & 0.390 & 0.409 & 0.023 \\ 
                    materials & 18.267 & 0.804 & 0.177 & 0.336 & 0.327 & 0.331 & 0.022 \\ 
                    mic & 21.967 & 0.905 & 0.089 & 0.598 & 0.555 & 0.576 & 0.013 \\ 
                    ship & 11.640 & 0.56 & 0.423 & 0.064 & 0.059 & 0.061 & 0.184 \\ 
                    \hline 
                    Avg. & 18.684 & 0.806 & 0.189 & 0.377 & 0.376 & 0.372 & 0.048 \\ 
                    \hline
                    \hline
            \end{tabular}
            }
           \caption{FVS, Discrete (local), 10 views}
           \label{tab:supp:nerf_fvs_10}
    \end{subtable}

    \begin{subtable}[h]{1\columnwidth}
        \centering
            \resizebox{1\columnwidth}{!}{%
            \begin{tabular}{ l |  c  c c | c c c c } 
                    Scene &
                    PSNR $\uparrow$ & 
                    SSIM $\uparrow$ & 
                    LPIPS $\downarrow$ & 
                    Accu. $\uparrow$ & 
                    Comp. $\uparrow$ & 
                    F1-Score $\uparrow$ & 
                    Chamfer $\downarrow$ \\
                    \hline
                    \hline
                    chair & 18.633 & 0.846 & 0.165 & 0.365 & 0.240 & 0.290 & 0.090 \\ 
                    drums & 14.700 & 0.756 & 0.261 & 0.356 & 0.386 & 0.371 & 0.029 \\ 
                    ficus & 19.133 & 0.854 & 0.133 & 0.413 & 0.681 & 0.514 & 0.017 \\ 
                    hotdog & 15.570 & 0.817 & 0.254 & 0.179 & 0.138 & 0.156 & 0.098 \\ 
                    lego & 17.960 & 0.801 & 0.189 & 0.351 & 0.331 & 0.341 & 0.031 \\ 
                    materials & 17.870 & 0.798 & 0.186 & 0.248 & 0.401 & 0.306 & 0.103 \\ 
                    mic & 21.033 & 0.897 & 0.098 & 0.527 & 0.509 & 0.518 & 0.016 \\ 
                    ship & 10.050 & 0.545 & 0.431 & 0.079 & 0.099 & 0.088 & 0.176 \\ 
                    \hline
                    Avg. & 16.869 & 0.789 & 0.215 & 0.315 & 0.348 & 0.323 & 0.070 \\  
                    \hline
                    \hline
            \end{tabular}
            }
           \caption{Random, Discrete (local), 10 views}
           \label{tab:supp:nerf_random_10}
    \end{subtable}

     \caption{
        (Part A) Novel view synthesis and 3D reconstruction evaluation result of individual scenes in NeRF-Synthetic dataset.
     }
     \label{tab:individual_partA}
     \vspace{-5pt}
\end{table}

\begin{table}[t!]

    \begin{subtable}[h]{1\columnwidth}
        \centering
            \resizebox{1\columnwidth}{!}{%
            \begin{tabular}{ l |  c  c c | c c c c } 
                    Scene &
                    PSNR $\uparrow$ & 
                    SSIM $\uparrow$ & 
                    LPIPS $\downarrow$ & 
                    Accu. $\uparrow$ & 
                    Comp. $\uparrow$ & 
                    F1-Score $\uparrow$ & 
                    Chamfer $\downarrow$ \\
                    \hline
                    \hline
                    chair & 23.470 & 0.875 & 0.117 & - & - & - & - \\ 
                    drums & 17.750 & 0.789 & 0.198 & - & - & - & - \\ 
                    ficus & 20.070 & 0.87 & 0.105 & - & - & - & - \\ 
                    hotdog & 25.000 & 0.902 & 0.122 & - & - & - & - \\ 
                    lego & 22.700 & 0.838 & 0.136 & - & - & - & - \\ 
                    materials & 20.330 & 0.836 & 0.148 & - & - & - & - \\ 
                    mic & 25.500 & 0.941 & 0.055 & - & - & - & - \\ 
                    ship & 20.000 & 0.72 & 0.273 & - & - & - & - \\ 
                    \hline
                    Avg. & 21.853 & 0.846 & 0.144 & - & - & - & - \\ 
                    \hline
                    \hline
            \end{tabular}
            }
           \caption{Ours (Entropy), Discrete (free), 10 views}
           \label{tab:supp:nerf_entropy_10_free}
    \end{subtable}

    \begin{subtable}[h]{1\columnwidth}
        \centering
            \resizebox{1\columnwidth}{!}{%
            \begin{tabular}{ l |  c  c c | c c c c } 
                    Scene &
                    PSNR $\uparrow$ & 
                    SSIM $\uparrow$ & 
                    LPIPS $\downarrow$ & 
                    Accu. $\uparrow$ & 
                    Comp. $\uparrow$ & 
                    F1-Score $\uparrow$ & 
                    Chamfer $\downarrow$ \\
                    \hline
                    \hline
                    chair & 26.690 & 0.919 & 0.081 & - & - & - & - \\ 
                    drums & 20.200 & 0.847 & 0.154 & - & - & - & - \\ 
                    ficus & 21.520 & 0.892 & 0.095 & - & - & - & - \\ 
                    hotdog & 28.840 & 0.944 & 0.079 & - & - & - & - \\ 
                    lego & 25.220 & 0.893 & 0.099 & - & - & - & - \\ 
                    materials & 22.360 & 0.866 & 0.124 & - & - & - & - \\ 
                    mic & 27.480 & 0.954 & 0.046 & - & - & - & - \\ 
                    ship & 23.600 & 0.786 & 0.222 & - & - & - & - \\ 
                    \hline
                    Avg. & 24.489 & 0.888 & 0.113 & - & - & - & - \\ 
                    \hline
                    \hline
            \end{tabular}
            }
           \caption{Ours (Entropy), Discrete (free), 20 views}
           \label{tab:supp:nerf_entropy_20_free}
    \end{subtable}

    \begin{subtable}[h]{1\columnwidth}
        \centering
            \resizebox{1\columnwidth}{!}{%
            \begin{tabular}{ l |  c  c c | c c c c } 
                    Scene &
                    PSNR $\uparrow$ & 
                    SSIM $\uparrow$ & 
                    LPIPS $\downarrow$ & 
                    Accu. $\uparrow$ & 
                    Comp. $\uparrow$ & 
                    F1-Score $\uparrow$ & 
                    Chamfer $\downarrow$ \\
                    \hline
                    \hline
                    chair & 27.700 & 0.932 & 0.067 & 0.489 & 0.557 & 0.521 & 0.016 \\ 
                    drums & 18.730 & 0.822 & 0.179 & 0.512 & 0.561 & 0.536 & 0.014 \\ 
                    ficus & 22.800 & 0.905 & 0.083 & 0.744 & 0.867 & 0.801 & 0.008 \\ 
                    hotdog & 28.430 & 0.938 & 0.085 & 0.319 & 0.461 & 0.377 & 0.045 \\ 
                    lego & 24.730 & 0.885 & 0.105 & 0.553 & 0.534 & 0.543 & 0.012 \\ 
                    materials & 22.300 & 0.869 & 0.117 & 0.572 & 0.534 & 0.552 & 0.012 \\ 
                    mic & 25.800 & 0.942 & 0.055 & 0.670 & 0.707 & 0.688 & 0.009 \\ 
                    ship & 21.000 & 0.752 & 0.255 & 0.186 & 0.166 & 0.175 & 0.053 \\ 
                    \hline 
                    Avg. & 23.936 & 0.881 & 0.118 & 0.506 & 0.548 & 0.524 & 0.021 \\ 
                    \hline
                    \hline
            \end{tabular}
            }
           \caption{Ours (Entropy), Discrete (local), 20 views}
           \label{tab:supp:nerf_entropy_20}
    \end{subtable}

    \begin{subtable}[h]{1\columnwidth}
        \centering
            \resizebox{1\columnwidth}{!}{%
            \begin{tabular}{ l |  c  c c | c c c c } 
                    Scene &
                    PSNR $\uparrow$ & 
                    SSIM $\uparrow$ & 
                    LPIPS $\downarrow$ & 
                    Accu. $\uparrow$ & 
                    Comp. $\uparrow$ & 
                    F1-Score $\uparrow$ & 
                    Chamfer $\downarrow$ \\
                    \hline
                    \hline
                    chair & 28.100 & 0.934 & 0.065 & 0.449 & 0.497 & 0.472 & 0.019 \\ 
                    drums & 18.967 & 0.825 & 0.174 & 0.484 & 0.528 & 0.505 & 0.015 \\ 
                    ficus & 22.770 & 0.904 & 0.086 & 0.723 & 0.835 & 0.775 & 0.009 \\ 
                    hotdog & 29.830 & 0.947 & 0.075 & 0.408 & 0.425 & 0.416 & 0.017 \\ 
                    lego & 25.070 & 0.89 & 0.102 & 0.551 & 0.529 & 0.540 & 0.012 \\ 
                    materials & 21.870 & 0.858 & 0.13 & 0.477 & 0.487 & 0.482 & 0.013 \\ 
                    mic & 25.330 & 0.939 & 0.059 & 0.716 & 0.752 & 0.733 & 0.008 \\ 
                    ship & 15.830 & 0.679 & 0.327 & 0.132 & 0.122 & 0.127 & 0.094 \\ 
                    \hline 
                    Avg. & 23.471 & 0.872 & 0.127 & 0.493 & 0.522 & 0.506 & 0.024 \\ 
                    \hline
                    \hline
            \end{tabular}
            }
           \caption{FVS, Discrete (local), 20 views}
           \label{tab:supp:nerf_fvs_20}
    \end{subtable}

    \begin{subtable}[h]{1\columnwidth}
        \centering
            \resizebox{1\columnwidth}{!}{%
            \begin{tabular}{ l |  c  c c | c c c c } 
                    Scene &
                    PSNR $\uparrow$ & 
                    SSIM $\uparrow$ & 
                    LPIPS $\downarrow$ & 
                    Accu. $\uparrow$ & 
                    Comp. $\uparrow$ & 
                    F1-Score $\uparrow$ & 
                    Chamfer $\downarrow$ \\
                    \hline
                    \hline
                    chair & 21.667 & 0.871 & 0.127 & 0.387 & 0.331 & 0.357 & 0.045 \\ 
                    drums & 17.530 & 0.805 & 0.199 & 0.417 & 0.473 & 0.444 & 0.020 \\ 
                    ficus & 21.370 & 0.887 & 0.105 & 0.631 & 0.796 & 0.704 & 0.010 \\ 
                    hotdog & 21.270 & 0.884 & 0.151 & 0.358 & 0.337 & 0.347 & 0.030 \\ 
                    lego & 22.767 & 0.859 & 0.13 & 0.431 & 0.405 & 0.417 & 0.023 \\ 
                    materials & 20.267 & 0.838 & 0.145 & 0.486 & 0.435 & 0.459 & 0.015 \\ 
                    mic & 24.630 & 0.932 & 0.064 & 0.696 & 0.612 & 0.651 & 0.011 \\ 
                    ship & 13.980 & 0.634 & 0.357 & 0.163 & 0.137 & 0.149 & 0.065 \\ 
                    \hline
                    Avg. & 20.435 & 0.839 & 0.160 & 0.446 & 0.441 & 0.441 & 0.027 \\ 
                    \hline
                    \hline
            \end{tabular}
            }
           \caption{Random, Discrete (local), 20 views}
           \label{tab:supp:nerf_random_20}
    \end{subtable}

     \caption{
        (Part B) Novel view synthesis and 3D reconstruction evaluation result of individual scenes in NeRF-Synthetic dataset.
     }
     \label{tab:individual_partB}
     \vspace{-5pt}
\end{table}

\subsection{More results for termination entropy study}
In Sec. 5.3, we present a study about using termination entropy as the criterion for calculating information gain.
Here we present more details of the study with qualitative comparisons of individual scenes in the dataset, and quantitative comparisons of the scenes. 
The quantitative comparison is presented in \cref{tab:individual_partA} and \cref{tab:individual_partB} while the qualitative comparison are shown in \cref{fig:supp:nerf_discrete_local_10} and \cref{fig:supp:tt_discrete_local_50}.

For the quantitative result, termination entropy-based method performs consistently better than the baselines in all the scenes. 
The qualitative comparison shows that the entropy-based method allows the agent to explore more diverse viewpoints and builds better 3D models with limited number of viewpoints.

\subsection{More qualitative result}
We present a qualitative comparison between our proposed method and DVGO \cite{SunSC22dvgo} in \cref{fig:supp:nerf_ours}.
As same as most NeRF-based prior work \cite{mildenhall2021nerf, garbin2021fastnerf}, DVGO optimizes the scene representation based on a set of collected images in an \textit{offline training} manner. 
In contrast, we optimize the representation in an \textit{online} fashion by adding new observations into the training set incrementally.
The agent has to plan for the next best views and perceive new observations based on the current scene representation. 
Moreover, our method uses the images rendered by DVGO as the training images, which implies that DVGO serves as the ``ground-truth'' and the upper bound of our method. 

We also include a video showing the active 3D reconstruction process.










\newpage


\end{document}